\documentclass[letterpaper, 10 pt, conference]{ieeeconf}  %

\IEEEoverridecommandlockouts                              %

\usepackage{cite}
\usepackage{amsmath,amssymb,amsfonts}
\usepackage{algorithmic}
\usepackage{graphicx}
\usepackage{textcomp}
\usepackage[table]{xcolor}
\usepackage{url}
\usepackage{multirow}
\usepackage{pifont}
\usepackage{xparse}

\def\BibTeX{{\rm B\kern-.05em{\sc i\kern-.025em b}\kern-.08em
    T\kern-.1667em\lower.7ex\hbox{E}\kern-.125emX}}

\title{ \bf ASCENT: Transformer-Based Aircraft Trajectory Prediction in Non-Towered Terminal Airspace}

\author{Alexander Prutsch$^{1}$, David Schinagl$^{1}$ and Horst Possegger$^{1}$%
\thanks{$^{1}\,$Corresponding author: {\tt alexander.prutsch@tugraz.at}. All authors are with the Institute of Visual Computing, Graz University of Technology, Austria. 
This work was partially funded by the Austrian Research Promotion Agency (FFG) under the project SAFER (894164).}
}

\begin{document}

\def\mn{ASCENT}
\def\icaoone{KBTP}
\def\icaotwo{KAGC}
\newcommand{\cmark}{\ding{51}} %
\newcommand{\xmark}{\ding{55}} %
\newcommand{\hp}[1]{\textcolor{blue}{#1}}
\newcommand{\ap}[1]{\begingroup\color{black}#1\endgroup}
\newcommand{\apnew}[1]{\begingroup\color{black}#1\endgroup}
\definecolor{rcol}{rgb}{0.9,0.9,0.9}
\newcommand{\sv}[1]{\underline{#1}}
\newcommand{\bv}[1]{\textbf{#1}}
\newcommand{\ourrow}[0]{\rowcolor{rcol}\mn~(Ours)}

\def\eg{\emph{e.g.,}~} \def\Eg{\emph{E.g.,}~}
\def\ie{\emph{i.e.,}~} \def\Ie{\emph{I.e.,}}
\def\fe{\emph{f.e.,}~} \def\Fe{\emph{F.e.,}}
\def\cf{\emph{cf.}~} \def\Cf{\emph{Cf.}}
\def\etc{\emph{etc}\onedot} 
\def\vs{\emph{vs}\onedot}
\def\wrt{w.r.t\onedot~} 
\def\dof{d.o.f\onedot}
\def\etal{\emph{et al}.~}
\def\etc{\emph{etc}\dots~}

\maketitle
\thispagestyle{empty}
\pagestyle{empty}

\begin{abstract}
Accurate trajectory prediction can improve General Aviation safety in non-towered terminal airspace, where high traffic density increases accident risk.
We present \mbox{\mn}, a lightweight transformer-based model for multi-modal 3D aircraft trajectory forecasting, which integrates domain-aware 3D coordinate normalization and parameterized predictions.
\mbox{\mn} employs a transformer-based motion encoder and a query-based decoder, enabling the generation of diverse maneuver hypotheses with low latency.
Experiments on the TrajAir and TartanAviation datasets demonstrate that our model outperforms prior baselines, as the encoder effectively captures motion dynamics and the decoder aligns with structured aircraft traffic patterns.
Furthermore, ablation studies confirm the contributions of the decoder design, coordinate-frame modeling, and parameterized outputs.
These results establish \mn~as an effective approach for real-time aircraft trajectory prediction in non-towered terminal airspace.

\end{abstract}

\section{INTRODUCTION}
Accurate trajectory predictions are crucial for improving airspace safety systems, especially for General Aviation (GA) flights near non-towered airports.
GA includes all non-commercial civilian flights, such as recreational flying, business aviation, and aerial work (\eg agricultural aviation).
It represents the majority of the aircraft fleet; for example, around 90\% of all aircraft registered in the United States fall under this category~\cite{faa2025general}.
Commercial Aviation~(CA), which focuses on the transport of passengers or cargo, operates under stringent safety standards: flights operate in controlled airspace according to structured procedures, with departures and arrivals conducted at towered airports under the active management of air traffic control (ATC).
However, most airports are \emph{non-towered}; it is estimated that only about 4\% of the roughly 20,000 airports in the US have control towers~\cite{johnson2017estimating}.
With no ATC active, pilots must rely on standard procedures, for example Federal Aviation Administration~(FAA) \emph{Standard Airport Traffic Patterns}~\cite{faa2022airplane}, and their own judgment to maintain safety.
GA flights frequently operate between non-towered airports, thereby facing a \emph{higher risk of accidents}.
The accident rate in GA is approximately 30 times higher than in commercial aviation (CA)~\cite{ntsb2023dashboard}, with fatal accidents occurring predominantly in the vicinity of airports.

\begin{figure}[t]
    \centering
    \includegraphics[trim={0mm, 0cm, 0mm, 0mm}, clip, width=0.8\columnwidth]{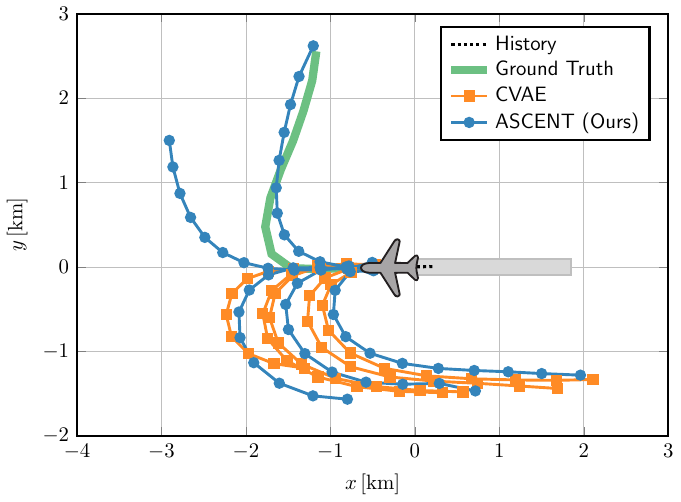}
    \vspace{-0.25cm}
    \caption{
    Comparison between a CVAE-based decoder, widely applied in aircraft trajectory prediction, and our \mn~model employing a query-based decoder.
    The scenario shows an aircraft take-off: our model, leveraging learnable mode queries, predicts a diverse set of trajectories that also capture the actual aircraft movement, while the baseline only predicts the common traffic pattern at the data-collection airport.
    }
    \label{fig:teaser1}
    \vspace{-0.5cm}
\end{figure}

The wide spread of GA, together with its relatively high accident rate, highlights the need for safety systems in non-towered airspace.
One promising approach is the prediction of future aircraft trajectories, which typically relies on historical aircraft movement data combined with global context information.
The predictions can help detect potential conflicts at an early stage~\cite{chen2020deep}, thereby improving GA safety in non-towered terminal areas.
Additionally, trajectory prediction can be applied in other tasks, \eg to enhance aircraft traffic flow~\cite{lin2019deep} and improve arrival time/delay prediction~\cite{wang2018hybrid}.
Previous research on aircraft trajectory prediction has introduced datasets from non-towered airspace~\cite{patrikar2022trajair, patrikar2025image} and developed deep learning-based methods~\cite{patrikar2022trajair, navarro2022social, yin2023context, yin2025aircraft, yang2025goodflight}.%

Analogously, trajectory prediction has become a key challenge in autonomous driving, where anticipating the future positions of surrounding traffic participants strongly influences self-driving vehicle decision-making and path planning.
Interest in autonomous driving has therefore driven a significant research focus on motion forecasting for self-driving vehicles~\cite{shi2022motion, nayakanti2023wayformer, wang2023prophnet, zhou2023query, lan2023sept, cheng2023forecast, shi2024mtr++, zhang2024demo, song2024realmotion}.
While methods that originate from the autonomous driving domain target a similar task, they \emph{cannot be directly applied to aircraft trajectory prediction} due to \emph{distinct challenges}: unlike road traffic, airplanes do not follow strict lane networks; only runways and airspace procedures define how aircraft approach and depart airports.
Moreover, aircraft operate in three-dimensional space, whereas bird’s-eye-view prediction is mostly sufficient for vehicles. 
Furthermore, trajectory prediction benchmarks for aviation~\cite{patrikar2022trajair, patrikar2025image} feature significantly longer scenarios and prediction horizons spanning several minutes compared to only a few seconds in the automotive domain.

Although recent advances have been made, aircraft trajectory prediction remains less extensively studied than its counterpart in autonomous driving.
In the self-driving domain, state-of-the-art methods typically employ transformer-based~\cite{vaswani2017attention} modules to encode historical motion data, and combine it with structured environmental information, \eg on lane topology~\cite{gao2020vectornet}.
Recent advances have been largely driven by improved coordinate modelings~\cite{zhou2023query}, decoder designs~\cite{zhou2024smartrefine, zhang2024demo} and refined processing schemes~\cite{zhou2024smartrefine, shi2024mtr++, song2024realmotion, zhang2024demo}.
At the same time, the field of aviation trajectory prediction~\cite{yin2025aircraft, yang2025goodflight} follows different approaches, particularly in terms of decoder architectures, highlighting a divergence between the two domains.
To advance trajectory prediction for General Aviation, we address the unique challenges of aircraft motion and propose a novel approach leveraging insights from state-of-the-art methods in the self-driving domain.
We propose an \emph{\underline{A}ircraft \underline{S}equen\underline{c}e \underline{En}coding \underline{T}ransformer}~(\mn) featuring a lightweight yet highly capable transformer-based architecture. %
In the decoder design, we diverge from common practices in aircraft trajectory prediction by using a mode query-based approach~\cite{cheng2023forecast}, which better suits the problem domain compared to diffusion or variational autoencoder-based modules of related work~\cite{yin2023context, yin2025aircraft, yang2025goodflight}.
As highlighted in Figure~\ref{fig:teaser1}, our approach generates more diverse outputs, thereby better covering the range of possible future actions.
We evaluate our approach on the widely used TrajAir dataset~\cite{patrikar2022trajair} and provide results on the recently proposed TartanAviation dataset~\cite{patrikar2025image}.

In summary, our main contributions include:
\begin{itemize}
    \item We propose a novel architecture for aircraft trajectory prediction, achieving a new state-of-the-art on the TrajAir dataset across multiple evaluation settings.
    \item To the best of our knowledge, we are the first to report trajectory prediction results on the TartanAviation dataset, including cross-dataset evaluations.
    \item We perform an extensive ablation study of our model demonstrating the effectiveness of our architecture.
\end{itemize}

\section{RELATED WORK}
\subsection{Aircraft Trajectory Prediction}
Patrikar~\etal~\cite{patrikar2022trajair} propose alongside their TrajAir dataset also a baseline method (TrajAirNet) for aircraft trajectory prediction in non-towered airspace.
Initially, the sequence of 3D positions of each agent, given in absolute coordinates, are encoded using temporal convolutional networks~\cite{lea2017temporal}, before interactions between all agents are modeled using graph attention networks~\cite{velivckovic2017graph}. 
The airplane trajectories are decoded using conditional variational autoencoders~\cite{yan2016attribute2image}.
To model the flight dynamics, the model predicts acceleration values which are then integrated to 3D positions using Verlet integration~\cite{verlet1967computer}.
Social-PatteRNN~\cite{navarro2022social} proposes a variational RNN~(VRNN)-based~\cite{chung2015recurrent} approach with a novel context modeling which is also evaluated on the TrajAir dataset.

\begin{figure*}[t]
    \centering
    \vspace{0.17cm}
    \includegraphics[width=0.93\linewidth]{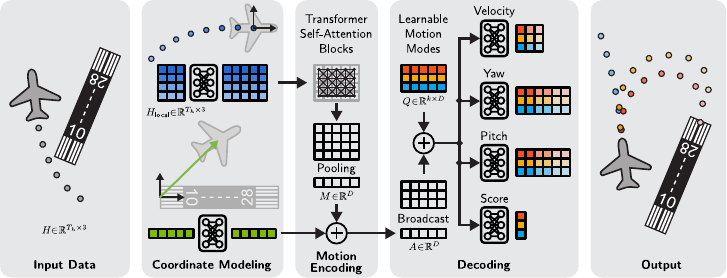}
    \vspace{-0.25cm}
    \caption{
    Overview of our \mn~architecture. For clarity of presentation, visualizations are reduced to 2D.
    First, we normalize the historical 3D flight path with respect to the current aircraft pose.
    Next, we encode this local history using attention blocks and combine it with global context through positional embeddings.
    To generate multi-modal future trajectories, we employ learnable mode queries that capture potential actions, such as landing or turning.
    }
    \label{fig:arch}
    \vspace{-0.5cm}
\end{figure*}

Also Yin~\etal\cite{yin2023context} focus on context modeling for trajectory prediction and propose a diffusion-based approach.
Evaluation is conducted on a custom, non-public dataset recorded at the towered Singapore Changi Airport.
Their model combines a Trajectron++-based~\cite{salzmann2020trajectron} motion encoder that operates on absolute coordinates with a separate context encoder.
The latter processes additional features such as landing time and target runway, which are only applicable in towered airspace.
The trajectory and context features are concatenated and future trajectories are predicted using a decoder inspired by MID~\cite{gu2022stochastic}, which models a reverse diffusion process.
The follow-up work ATP~\cite{yin2025aircraft} leverages a local history of similar past aircraft behaviors, which achieves strong results on TrajAir as well as the private Singapore Changi Airport dataset.
Recently, GooDFlight~\cite{yang2025goodflight} also proposes a diffusion-based model for flight trajectory prediction.
This approach adopts goal-oriented modeling: it first predicts goal coordinates, which are then used in a goal-guided decoder. GooDFlight currently achieves state-of-the-art results on the TrajAir dataset.

Despite extensive research on aircraft trajectory prediction, existing models differ substantially from state-of-the-art in autonomous driving.
Current approaches in aviation often employ decoders based on generative stochastic modeling, \ie variational methods~\cite{patrikar2022trajair, navarro2022social} or diffusion processes~\cite{yin2023context, yin2025aircraft, yang2025goodflight}, to predict multi-modal trajectories.
Generative approaches have shown particular strength in domains like pedestrian trajectory prediction~\cite{xu2022socialvae, rempe2023trace, capellera2025unified}, where agents move freely and unpredictably.
However, generative designs are not adopted in leading self-driving models~\cite{zhou2023query, zhou2024smartrefine, shi2024mtr++, zhang2024demo}.
Although GA aircraft do not strictly follow lanes as cars do, they still adhere to predefined traffic patterns, resulting in structured and discrete motion modes.
Given the maturity and demonstrated accuracy of trajectory prediction in autonomous driving, it is natural to ask whether insights and architectures from that field can be adapted to the aviation domain.
Such cross-domain transfer offers a promising direction for advancing aircraft trajectory prediction, despite the inherent differences between the two problem settings.

\subsection{Trajectory Prediction for Autonomous Driving}
Trajectory prediction for self-driving vehicles is a well-studied task.
Large-scale benchmark datasets like the Waymo Open Motion Dataset~\cite{ettinger2021large} and nuScenes~\cite{caesar2020nuscenes} have significantly driven advances in motion prediction modeling.
In contrast to airspace trajectory prediction, vehicle movements are heavily restricted by the lane topology and other environmental context like traffic lights and lane markings.
In addition, methods mostly operate only in a 2D bird's eye view representation.
Motion prediction methods for autonomous driving commonly follow the same high-level architecture, where first historical agent movements and map elements are encoded in a local coordinate system, before global positional embeddings are employed and the relationships between scene elements are learned.
After initial works explored CNNs~\cite{chai2020multipath} or LSTMs~\cite{mercat2020multi}, attention blocks~\cite{vaswani2017attention} have emerged as a popular approach for encoding agent motion behavior~\cite{shi2022motion, cheng2023forecast}.
Scene encoding is commonly done using graph neural networks~\cite{gao2020vectornet, liang2020learning, zhou2023query, jia2023hdgt, cui2023gorela} or transformers~\cite{liu2021multimodal, shi2022motion, nayakanti2023wayformer, cheng2023forecast}. %
Global positional embeddings can be implemented using MLP-based~\cite{cheng2023forecast}, sinusoidal~\cite{shi2022motion}, or Fourier-space embeddings~\cite{zhou2023query}.
To decode multi-modal trajectories, the common practice is to use learnable mode queries which model different potential future agent directions.
The mode queries can either be used in simple MLP-based decoders~\cite{cheng2023forecast, prutsch24efficient} or as queries in a detection transformer~(DETR)-like decoder~\cite{shi2022motion, shi2024mtr++}.

Methods developed for autonomous driving cannot be directly applied to aircraft trajectory prediction, as they rely heavily on lane topologies for guidance and are restricted to a 2D space.
They are also tailored to much denser traffic scenarios but typically operate only over short time horizons of a few seconds -- insufficient for aviation needs.
In addition, these models are often large and require extensive, diverse training datasets.
In contrast to terminal airspace, where the environment is static, methods for self-driving vehicles must generalize across diverse and dynamically evolving road networks.
Nevertheless, concepts from autonomous driving like local motion encoding with global positional embeddings and mode query-based decoders represent promising directions for advancing aircraft trajectory prediction.

\section{Aircraft Sequence Encoding Transformer}
In this section, we formally define the aircraft trajectory prediction problem and introduce our novel \mn~architecture (see Figure~\ref{fig:arch} for an overview), which integrates best practices from autonomous driving research and extends them to the unique challenges of 3D flight paths.
We first describe our input coordinate modeling, where positional and angular normalization are applied to effectively capture motion patterns from historical trajectories.
To preserve global context after normalization, we introduce a novel 3D positional embedding. Both normalization and embedding must account for the horizontal (yaw) and vertical (pitch) orientation of aircraft, unlike vehicles that typically vary only in yaw.
We then outline our motion pattern encoding.
Finally, we detail our decoder design, where learnable mode queries and flight parameter outputs are employed to better align the architecture with the characteristics of the aviation domain.

\subsection{Problem Statement}
Aircraft trajectory prediction is the task of forecasting the 3D positions of an aircraft over $T_f$ future steps, denoted as $F_{GT} \in \mathbb{R}^{T_f \times 3}$, given its historical motion over $T_h$ steps ($H \in \mathbb{R}^{T_h \times 3}$).
Since future motion can correspond to different maneuvers, \eg landing or turning, models typically predict a set of $k$ trajectory candidates, resulting in multi-modal outputs $F \in \mathbb{R}^{k \times T_f \times 3}$ with associated probability scores $S \in \mathbb{R}^{k}$.
In non-towered terminal airspace, the historical data $H$ is typically referenced to a static Automatic Dependent Surveillance–Broadcast~(ADS-B) receiver.
Additionally, auxiliary context such as weather conditions or the trajectories of neighboring aircraft can be incorporated.

\begin{figure*}[t]
    \vspace{0.17cm}
    \centering
    \includegraphics[trim={0.5cm, 0.5cm, 2cm, 3.5cm}, clip, width=0.32\linewidth]{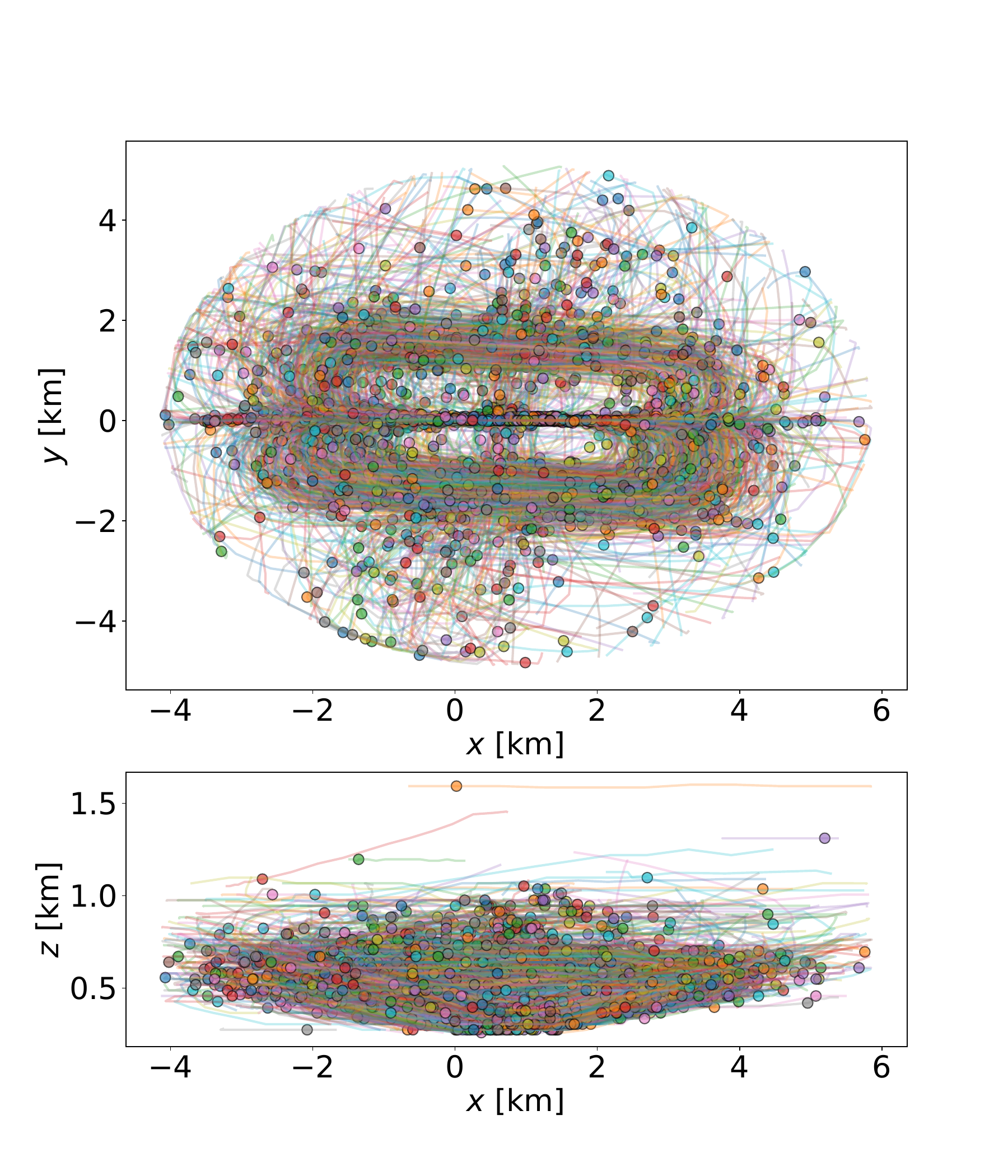}
    \hspace{0.05cm}
    \includegraphics[trim={0.5cm, 0.5cm, 2cm, 3.5cm}, clip, width=0.32\linewidth]{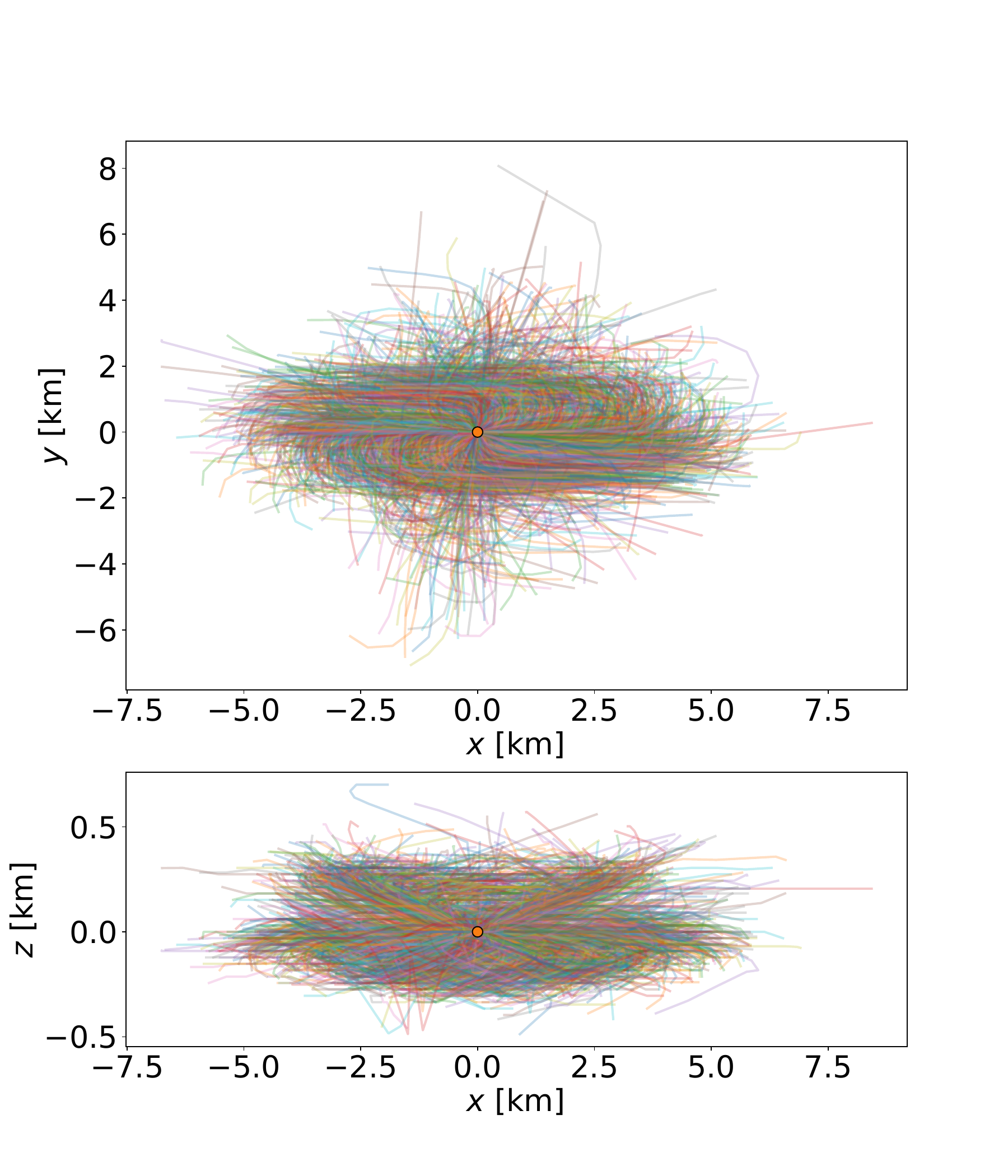}
    \hspace{0.05cm}
    \includegraphics[trim={0.5cm, 0.5cm, 2cm, 3.5cm}, clip, width=0.32\linewidth]{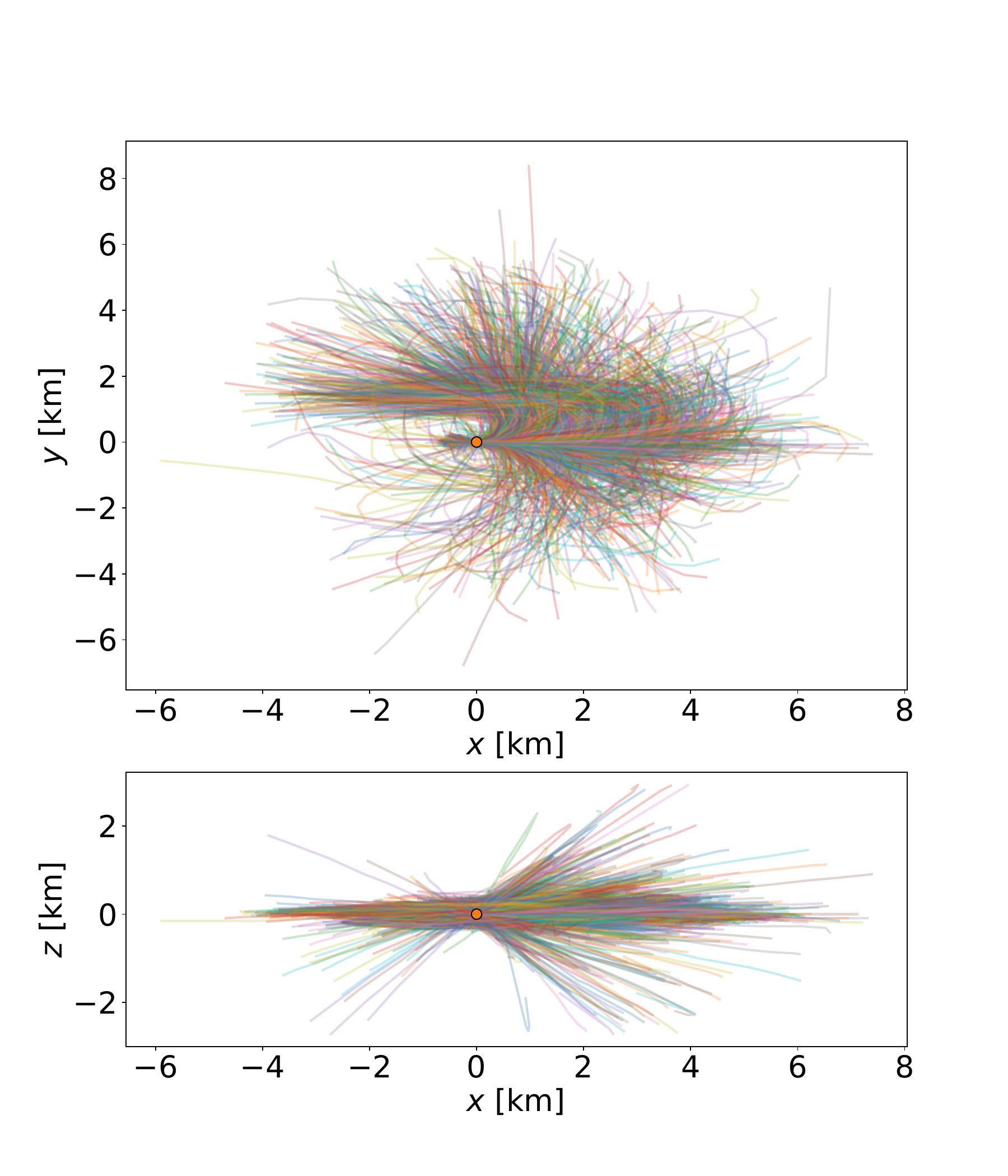}
    \vspace{-0.5cm}
    \caption{
    Comparison of different coordinate system normalization approaches on trajectories from the TrajAir dataset~\cite{patrikar2022trajair}.
    We show the current aircraft position (circular markers), along with historical (11\,s) and future (120\,s) trajectories.
    We compare global coordinates (left), object-centered normalization (middle), and our combined positional–angular normalization (right).
    For each method, the bird’s-eye view is shown on top, with vertical trajectories below.
    The global coordinate visualization shows aircraft approaching and departing from the runway, which is aligned with the x-axis.
    Normalizing with respect to the aircraft position moves all centers to the origin, while angular normalization aligns orientations, producing structured clusters of similar maneuvers, such as left and right turns.
    The top-right plot highlights a higher frequency of left turns, reflecting the left-traffic pattern at the \icaoone~airport.
    }
    \label{fig:coord}
    \vspace{-0.5cm}
\end{figure*}

\subsection{Model Architecture}

\subsubsection{Coordinate Modeling}
Generally, the future path of an aircraft depends on its historical motion pattern and the global position context, \eg distance and orientation to the runway.
Aircraft trajectory prediction datasets~\cite{patrikar2022trajair, patrikar2025image} are commonly recorded using a static scene coordinate system.
By directly using the global coordinates it is possible to model both, however, this can lead to a global position bias and hinders effectively learning local motion patterns, such as turns, dives or climb flights.
To circumvent this, it is a common practice to normalize each agent path with respect to the current agent position to obtain an agent-centric local coordinate system.
Naively, this can be either done by just shifting the trajectory by the object center, or more sophisticated by also aligning the orientation of the agent with the coordinate axis \eg aircraft looking in positive y-direction.
Figure~\ref{fig:coord} shows a comparison of global and normalized coordinates using example data from the TrajAir dataset.
In our approach, we apply positional and angular normalization to obtain $H_{\text{local}}$ which can be used to effectively learn motion patterns.
We estimate the orientation of the aircraft using the last and second-to-last position step.
Following, we compute the horizontal (yaw angle~$\gamma$) and the vertical (pitch angle~$\theta$) orientation using the arctangent.
Normalization is then done by subtracting the most-recent aircraft position and rotation around $\gamma$ and $\theta$. 

Using only agent-centric local coordinates neglects global information, \eg the relationship between the aircraft pose and a runway.
A common practice in the automotive domain to preserve global context is the use of positional embeddings.
We employ a novel 3D positional embedding using the 3D position $(x, y, z)$ and the orientation angles $(\gamma$ and $\theta)$, which we compute during normalization.

\begin{table*}[t]
\setlength{\tabcolsep}{3.5pt}
    \vspace{0.17cm}
    \caption{
    Trajectory prediction results on the dedicated TrajAir splits, along with average results across all four splits.
    All models predict $K{=}5$ trajectories.  
    The upper group follows the experiment setup of~\cite{patrikar2022trajair} (using $11\,\text{s}$ input).
    The lower group compares to results from~\cite{yin2025aircraft}, using their setup ($16\,\text{s}$ input). 
    Models are sorted by average minFDE$_5$. Errors are reported in km.
    }
    \vspace{-0.4cm}
    \label{tab:res_trajair1}
    \begin{center}
    \begin{tabular}{l||cc|cc|cc|cc||cc}
    \multicolumn{1}{c||}{\emph{Input: 11 s}} & \multicolumn{2}{c|}{7Days-1} & \multicolumn{2}{c|}{7Days-2} & \multicolumn{2}{c|}{7Days-3} & \multicolumn{2}{c||}{7Days-4} & \multicolumn{2}{c}{7Days-Avg} \\
    \multicolumn{1}{c||}{Method} & minADE$_5$ & minFDE$_5$ & minADE$_5$ & minFDE$_5$ & minADE$_5$ & minFDE$_5$ & minADE$_5$ & minFDE$_5$ & minADE$_5$ & minFDE$_5$ \\ \hline 
    Constant Velocity                               & 1.79 & 4.08 & 1.90 & 4.31 & 1.92 & 4.30 & 1.82 & 4.16 & 1.86 & 4.21 \\ %
    TransformerTF~\cite{giuliari2021transformer}    & 1.58 & 3.85 & 1.69 & 4.10 & 1.97 & 4.36 & 1.79 & 4.19 & 1.76 & 4.13 \\ %
    STG-CNN~\cite{mohamed2020social}                & 1.19 & 2.35 & 1.36 & 2.70 & 1.33 & 2.67 & 1.17 & 2.29 & 1.26 & 2.50 \\ %
    Nearest Neighbor                                & 3.13 & 2.70 & 1.92 & 1.99 & 3.41 & 2.69 & 2.59 & 2.58 & 2.76 & 2.49 \\ %
    TrajAirNet~\cite{patrikar2022trajair}           & \sv{0.73} & \sv{1.42} & \sv{0.81} & \sv{1.63} & \sv{0.86} & \sv{1.72} & \sv{0.71} & \sv{1.41} & \sv{0.78} & \sv{1.55} \\ %
    \ourrow                                         & \bv{0.33} & \bv{0.57} & \bv{0.37} & \bv{0.62} & \bv{0.36} & \bv{0.59} & \bv{0.33} & \bv{0.56} & \bv{0.35} & \bv{0.58} \\
    \multicolumn{2}{c}{} \\  [-0.2cm]
    \multicolumn{1}{c||}{\emph{Input: 16 s}} & \multicolumn{2}{c|}{7Days-1} & \multicolumn{2}{c|}{7Days-2} & \multicolumn{2}{c|}{7Days-3} & \multicolumn{2}{c||}{7Days-4} & \multicolumn{2}{c}{7Days-Avg} \\
    \multicolumn{1}{c||}{Method} & minADE$_5$ & minFDE$_5$ & minADE$_5$ & minFDE$_5$ & minADE$_5$ & minFDE$_5$ & minADE$_5$ & minFDE$_5$ & minADE$_5$ & minFDE$_5$ \\ \hline 
    Constant Velocity                               & 1.68 & 4.03 & 1.71 & 4.10 & 1.76 & 4.19 & 1.65 & 4.00 & 1.70 & 4.08 \\ %
    Nearest Neighbor                                & 0.86 & 1.69 & 0.90 & 1.79 & 1.01 & 1.93 & 0.83 & 1.67 & 0.90 & 1.77 \\ %
    TrajAirNet~\cite{patrikar2022trajair}           & 0.70 & 1.37 & 0.75 & 1.48 & 0.84 & 1.62 & 0.69 & 1.40 & 0.75 & 1.47 \\ %
    MID~\cite{gu2022stochastic}                     & 0.61 & 1.13 & 0.61 & 1.12 & 0.63 & 1.18 & 0.58 & 1.11 & 0.61 & 1.13 \\ %
    Expert-Traj~\cite{zhao2021expert}               & 0.57 & 1.05 & 0.56 & 1.03 & 0.63 & 1.15 & 0.53 & 0.99 & 0.57 & 1.06 \\ %
    LBA~\cite{zhong2022aware}                       & \sv{0.48} & 0.88 & 0.51 & 0.91 & 0.56 & 1.04 & 0.46 & 0.86 & 0.50 & 0.92 \\ %
    ATP~\cite{yin2025aircraft}                      & \sv{0.48} & \sv{0.85} & \sv{0.50} & 0.91 & 0.54 & 0.98 & \sv{0.45} & \sv{0.82} & \sv{0.49} & 0.89 \\ %
    CoATP w/ Abs. Loc.~\cite{yin2023context}                     & 0.50 & 0.88 & \sv{0.50} & \sv{0.89} & \sv{0.52} & \sv{0.92} & 0.47 & 0.83 & 0.50 & \sv{0.88} \\ %
    \ourrow                                         & \bv{0.31} & \bv{0.56} & \bv{0.35} & \bv{0.61} & \bv{0.34} & \bv{0.61} & \bv{0.33} & \bv{0.57} & \bv{0.33} & \bv{0.59}  \\
    \end{tabular}
    \end{center}
    \vspace{-0.5cm}
\setlength{\tabcolsep}{4pt}
\end{table*}

\subsubsection{Motion Encoding}
After normalization to local coordinates, we encode the historical aircraft path $H_{\text{local}} \in \mathbb{R}^{T_h \times 3}$ using self-attention blocks~\cite{vaswani2017attention}.
Before applying the transformer blocks, we first project $H_{\text{local}}$ to our feature space using a linear layer~\cite{lan2023sept} and add temporal encodings: we encode a numerical time index to a temporal embedding $\mathbb{R}^{D}$ using a linear layer.
After self-attention we apply max-pooling to reduce the dimensionality~\cite{lan2023sept}.
As a result we obtain a single encoded motion feature vector $M \in \mathbb{R}^{D}$ for an agent.
To model the global positional context, we then encode the pose ($x, y, z$, $\gamma$, $\theta$) using an MLP and add it to $M$ to get our feature vector $A \in \mathbb{R}^{D}$.

\subsubsection{Decoding with Learnable Mode Queries}
\mbox{\mn}~employs a simple multilayer perceptron~(MLP)-based decoder with learnable queries. 
This design has been proven to be a simple yet effective method for generating multi-modal trajectory outputs in the automotive domain~\cite{cheng2023forecast, prutsch24efficient}.
We take the encoded agent feature $A$, broadcast it to $\mathbb{R}^{k \times D}$ and add $k$ learnable mode queries $Q \in \mathbb{R}^{k \times D}$.
Following, we decode probability scores $S \in \mathbb{R}^{k}$ using a MLP.
To obtain the future positions $F \in \mathbb{R}^{k \times T_f \times 3}$ we do not directly predict the 3D positions, but rather predict flight parameters which we then use to compute the future positions.
Using kinematics instead of direct positional outputs allows the model to more easily capture the motion patterns of real-world actors~\cite{cui2020deep}.
Our network predicts the yaw and pitch angles of the aircraft and the flying speed.
The speed is predicted using a MLP with linear outputs.
To model the angular components, our model first predicts the sine and cosine of the target angle, from which the angle is then computed via the arctangent.
Using this, we wrap the angles to $[-\pi, \pi]$ while also achieving a stable loss behavior.
After predicting the individual components, we first compute the 3D positions based on the individual flight parameters.
Following, we apply a coordinate transformation inverse to our normalization, to transfer the predictions back to the global 3D space.

\section{EXPERIMENTAL SETUP}
\subsection{Datasets and Data Preprocessing}
We evaluate our method on two datasets, TrajAir~\cite{patrikar2022trajair} and TartanAviation~\cite{patrikar2025image}, both of which contain 3D aircraft trajectory data recorded via Automatic Dependent Surveillance–Broadcast (ADS-B).
The TrajAir dataset~\cite{patrikar2022trajair} was recorded at the non-towered Pittsburgh-Butler Regional Airport~(ICAO: KBTP) across 111 days between September 2020 and April 2021.
The TartanAviation~\cite{patrikar2025image} extends the ADS-B data from the TrajAir dataset with additional 222 days of recordings from the Pittsburgh-Butler Regional Airport.
The dataset also includes data from a second airport, the towered Allegheny County Airport (ICAO:~\icaotwo).
It also contains additional data (images and ATC voice), which are not relevant for our trajectory forecasting approach.

Regarding the TrajAir dataset~\cite{patrikar2022trajair} we follow the official implementation\footnote{\url{https://github.com/castacks/trajairnet}} and utilize the preprocessed data provided by the authors.
Their preprocessing pipeline removes corrupted and duplicated data points, filters out aircraft flying above 6000 feet MSL (mean sea level) or beyond a 5 kilometer radius away from the endpoints of the runway.
It also excludes empty sequences where no aircraft is recorded.
The framework provides training and validation sets for five distinct data splits: four splits each containing one week of data (\emph{7Days-{1-4}}) and one comprehensive split containing all available data (\emph{111Days}).
Analogously, we also use the official implementation\footnote{\url{https://github.com/castacks/TartanAviation/}} and preprocessed data for the TartanAviation dataset~\cite{patrikar2025image}.
Since no 5\,km radius filtering is applied, the dataset covers a wider range from the \icaoone~airport compared to TrajAir.
Since there are no official data splits for TartanAviation, we employ a custom split for each airport based on the processed sequence files: We take the first and last 25\% of files as one split (named S1) and the middle 50\% as a second split (S2). This reduces the impact of seasonal bias, with S1 covering potentially different seasonal conditions and S2 representing a more uniform set.

\subsection{Implementation Details}
We set the feature dimension $D$ in our model to 128 and use two self-attention blocks in our encoder.
In the decoder, each head (speed, yaw angle, pitch angle, and probability score) uses a 3-layer MLP with $2\cdot D$ units in the hidden layers.
The output layers are set to $T_f$ for the velocity head and to $2 \cdot T_f$ in the angular heads, where $T_f$ is the number of future steps.
The output layer for the probability head is a single value per trajectory.
The positional embeddings are implemented using a shallow MLP including two layers with dimension $D$.
We model the positional embedding of each actor based on its position $x, y, z$, yaw $\gamma$, and pitch $\theta$ as ($x, y, z, \sin \gamma, \cos \gamma, \sin \theta, \cos \theta$) to wrap angles.

We train our model on a single NVIDIA V100 GPU for 20 epochs using a batch size of 32 and an initial learning rate of 0.001, which is halved after 10 and 15 epochs. 
Loss computation is done using a winner-takes-all strategy: only the candidate trajectory that best fits the ground truth is considered for optimization (lowest $L_2$ distance).
We employ a smooth L1 loss~\cite{huber1964robust} as regression loss and use a standard cross-entropy loss as classification loss, \ie the best-fitting trajectory is assigned the highest probability score.

\subsection{Metrics}
We evaluate using the standard aircraft trajectory prediction metrics~\cite{patrikar2022trajair, yang2025goodflight}: the average displacement error~(ADE$_k$) and final displacement error~(FDE$_k$) reported in kilometers.
ADE is the mean displacement ($L_2$ distance) between prediction and ground truth trajectory across all future timesteps, whereas the FDE only takes the endpoints into account.
Following the standard protocol, both metrics are reported using the best-fitting prediction out of the $k$ candidates.

\begin{figure*}[t]
    \vspace{0.17cm}
    \centering
    \includegraphics[trim={0cm, 0cm, 0cm, 0.3cm}, clip, width=0.31\linewidth]{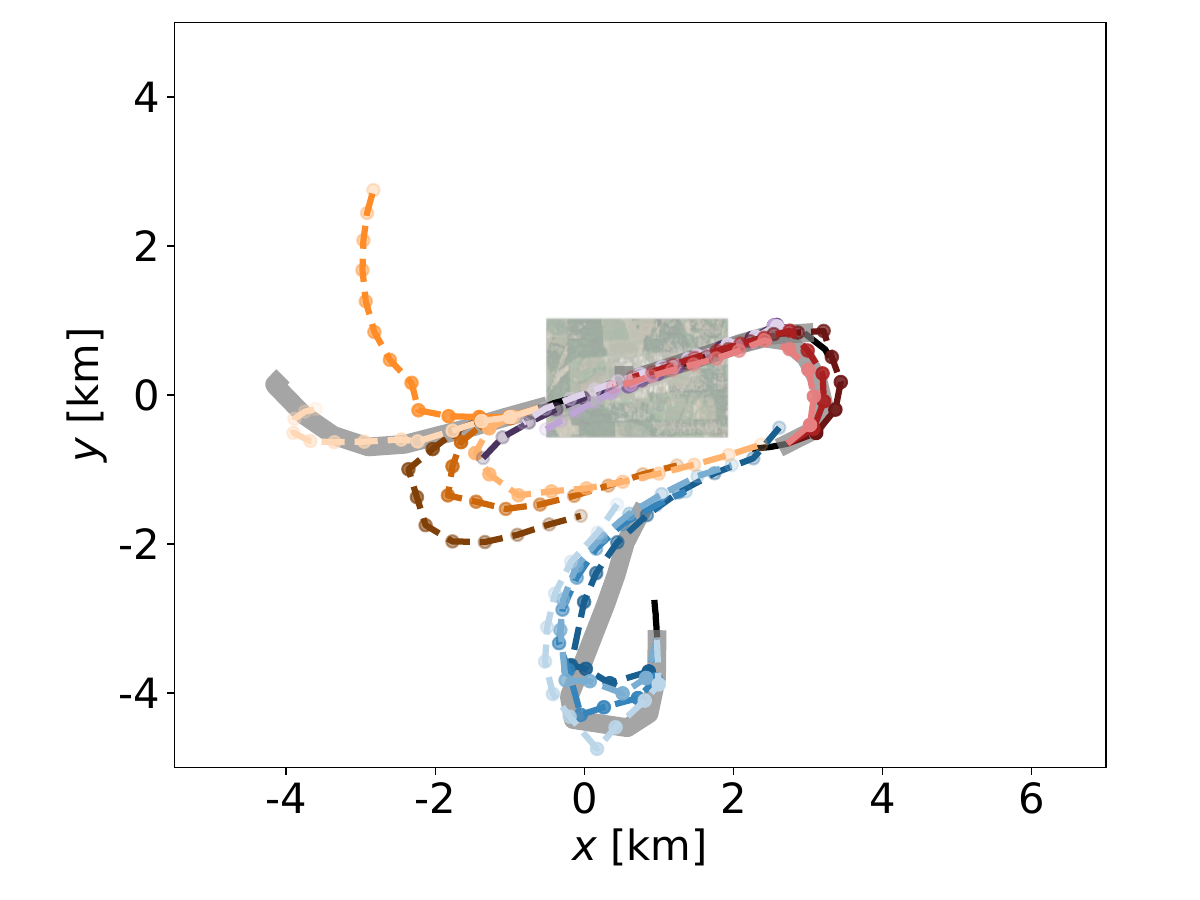}
    \hspace{0.05cm}
    \includegraphics[trim={0cm, 0cm, 0cm, 0.3cm}, clip, width=0.31\linewidth]{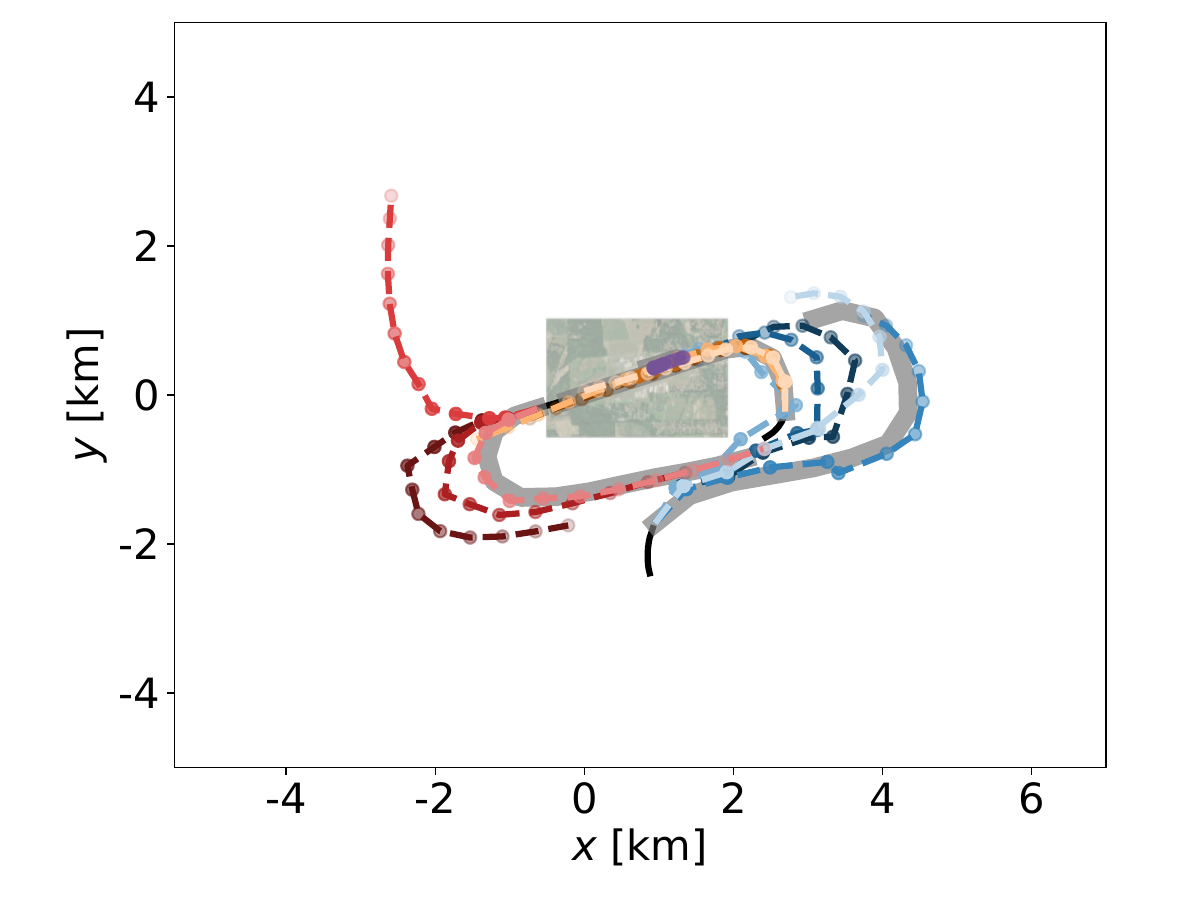}
    \hspace{0.05cm}
    \includegraphics[trim={0cm, 0cm, 0cm, 0.3cm}, clip, width=0.31\linewidth]{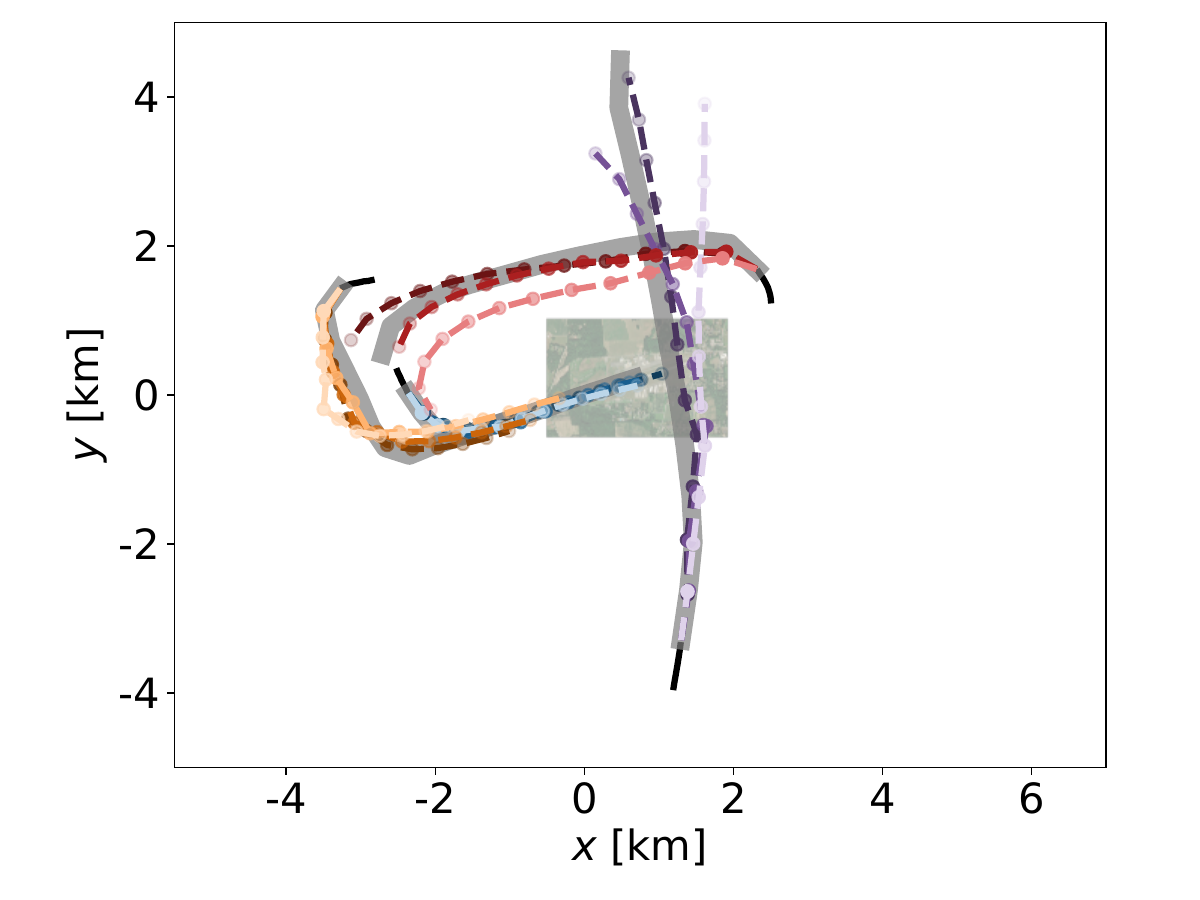} \\
    \vspace{-0.1cm}
    \hspace{0.005cm}
    \includegraphics[trim={0cm, 0cm, 0cm, 0.3cm}, clip, width=0.31\linewidth]{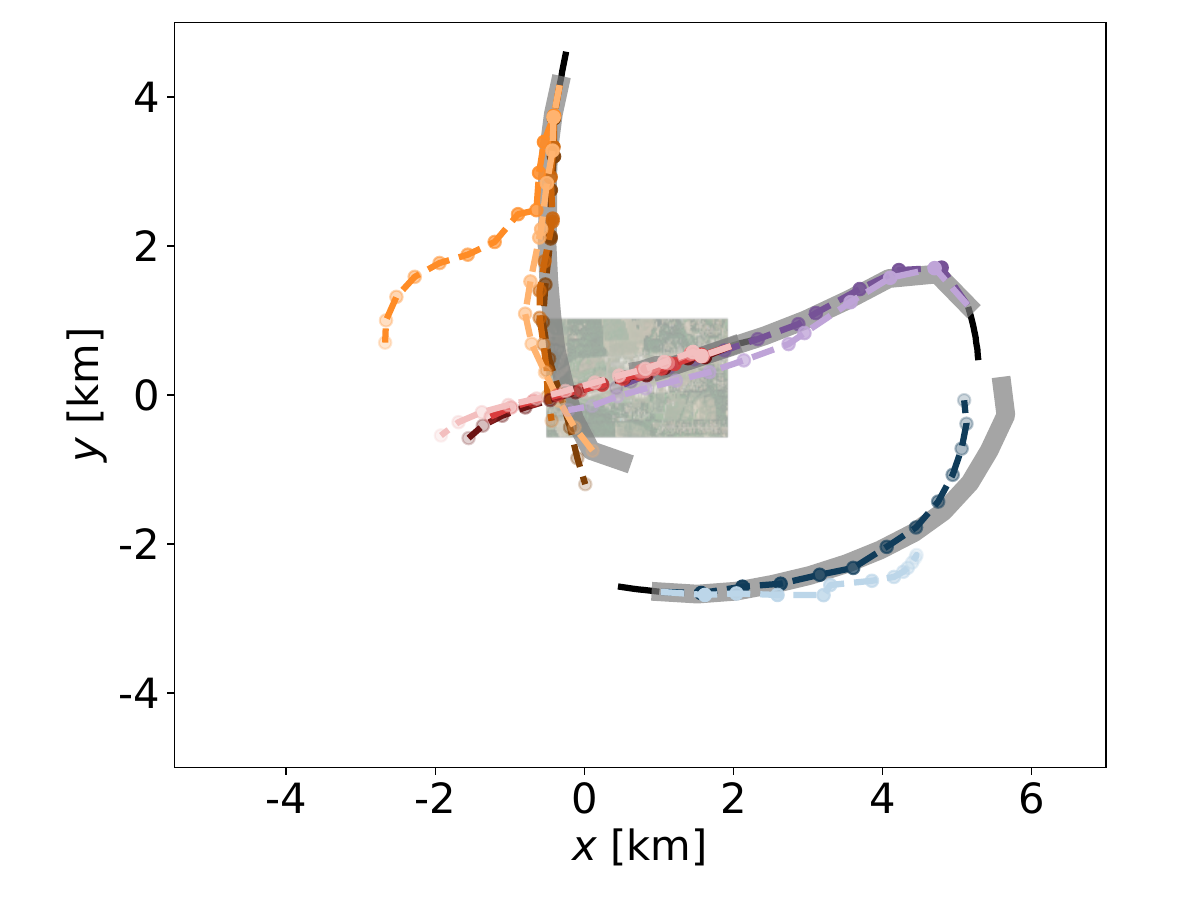}
    \hspace{0.05cm}
    \includegraphics[trim={0cm, 0cm, 0cm, 0.3cm}, clip, width=0.31\linewidth]{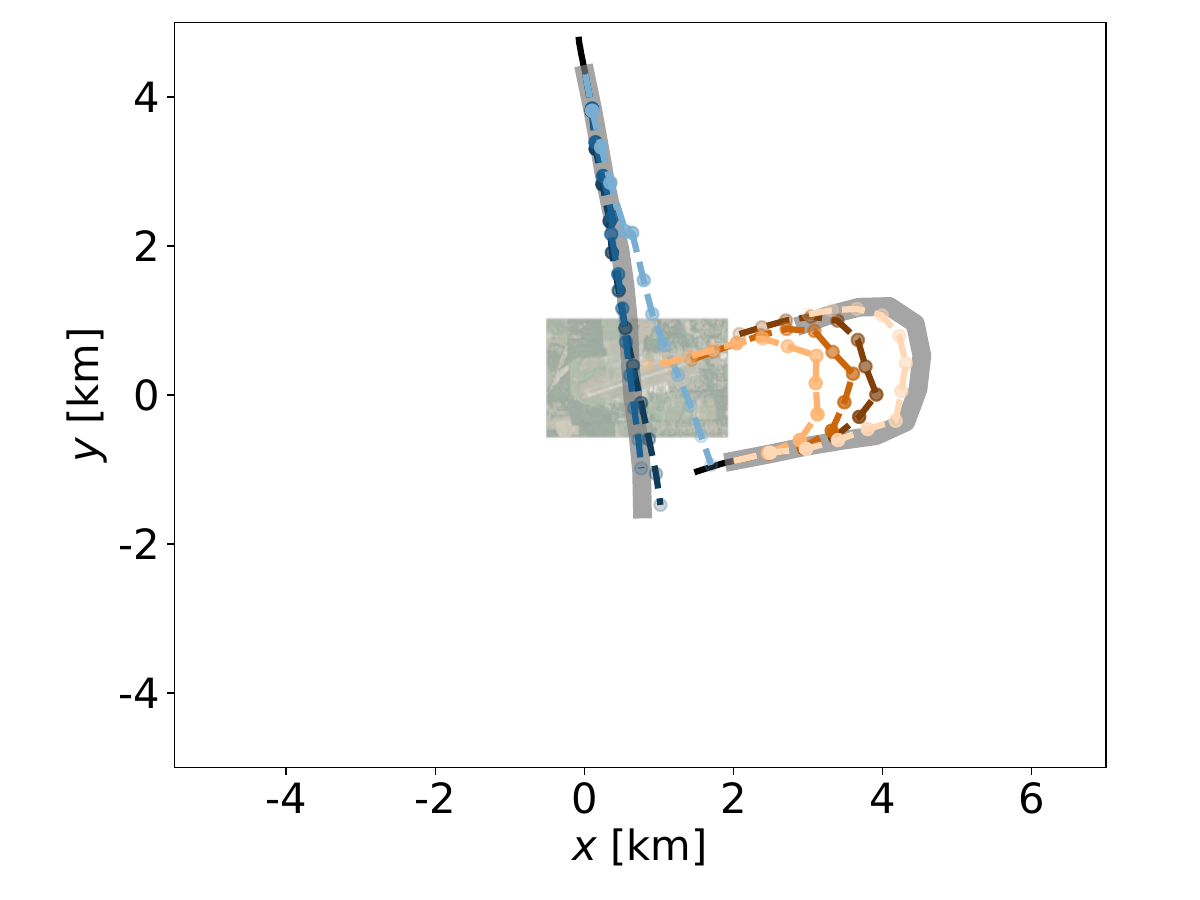}
    \hspace{0.05cm}
    \includegraphics[trim={0cm, 0cm, 0cm, 0.3cm}, clip, width=0.31\linewidth]{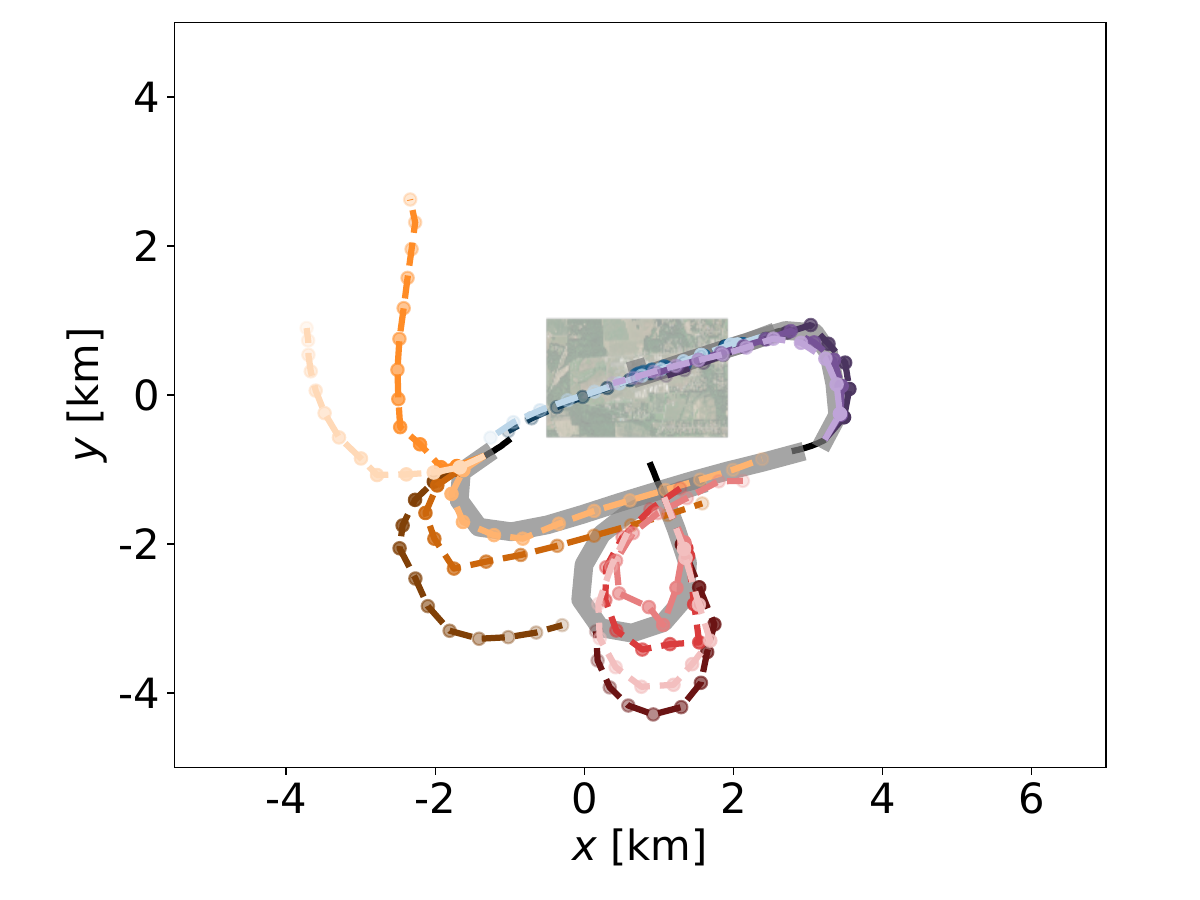} 
    \vspace{-0.4cm}
    \caption{
    Qualitative results on scenarios from the TrajAir~\cite{patrikar2022trajair} \emph{111Days} validation set. Visualizations show 11\,s of historical aircraft movement (black), 120\,s of ground truth future (gray) and predictions from our model (colored). For clearer visualization we exclude predictions with scores below 0.1.
    }
    \label{fig:res}
    \vspace{-0.5cm}
\end{figure*}

\section{RESULTS AND DISCUSSION}

We present our main results for trajectory prediction on the TrajAir~\cite{patrikar2022trajair} dataset in Tables~\ref{tab:res_trajair1}~and~\ref{tab:res_trajair2}.
Related work~\cite{patrikar2022trajair, yin2025aircraft, yang2025goodflight} report results using different historical data horizons as input and different numbers of output trajectory modes.
To provide an extensive evaluation and ensure comparability to state-of-the-art methods, we provide results for our approach on all three experiment configurations.
In Table~\ref{tab:res_trajair1} we compare our approach on the four dedicated one week data splits from the TrajAir dataset.
We follow the standard setup from the dataset paper~\cite{patrikar2022trajair}, 
using an 11\,s history (sampled at 1\,Hz) to predict the subsequent 120\,s at 0.1\,Hz, 
as well as the setting from related work~\cite{yin2025aircraft}, which uses a 16\,s history 
and a 120\,s prediction horizon, both sampled at 0.2\,Hz.
Our method significantly outperforms related work in both settings across all data splits.
Comparing both experimental settings shows that the advantage of using a larger historical context is offset by the sparser frame rate.
In Table~\ref{tab:res_trajair2} we evaluate our approach using the experiment settings presented in~\cite{yang2025goodflight}, which feature an extended history of 40\,s and also allow to predict 20 trajectory candidates, which naturally leads to lower overall errors.
Evaluation is done on the full TrajAir dataset (\emph{111Days} split) using the standard prediction horizon of 120\,s.
Again, our approach achieves favorable results compared to all methods.
The results highlight that \mn~significantly improves the state-of-the-art on TrajAir across all experiment settings used by related work.
Figure~\ref{fig:res} shows qualitative results for scenarios taken from the TrajAir dataset \emph{111Days} test set. 
We show the predictions of our \mn~model for all aircraft present in a scene, as well as the historical data and ground truth future.
The visualizations highlight that our model yields accurate trajectory outputs while also capturing different motion modes.
Table~\ref{tab:res_tartan} provides results on the TartanAviation dataset, where we evaluate a constant velocity baseline model and our approach.
We present results for training and testing on both splits, results marked with S1 mean testing is done on split S1 and vice-versa.
Our model achieves strong results on the \icaoone~airport, as well as, on the data from the towered \icaotwo~airport.

\begin{table}[t]
\setlength{\tabcolsep}{6pt}
    \vspace{0.17cm}
    \caption{
    Prediction results on the full TrajAir test set (\emph{111Days}).
    Upper group: following the setup of~\cite{yang2025goodflight}, all models use $h{=}40\,\text{s}$ input and predict $K{=}20$ trajectories. 
    Lower group: additional results for our model using the standard setup from the dataset paper~\cite{patrikar2022trajair}.
    Errors reported in km.
    }
    \vspace{-0.4cm}
    \label{tab:res_trajair2}
    \begin{center}
    \begin{tabular}{l||cc}
    \multicolumn{1}{c||}{\emph{Input: 40 s}} & \multicolumn{2}{c}{111Days} \\
    \multicolumn{1}{c||}{Method} & minADE$_{20}$ & minFDE$_{20}$ \\ \hline 
    Constant Velocity                               & 1.85 & 4.16 \\ %
    TransformerTF~\cite{giuliari2021transformer}    & 1.67 & 3.94 \\ %
    Nearest Neighbor                                & 1.97 & 2.70 \\ %
    STG-CNN~\cite{mohamed2020social}                & 1.37 & 2.35 \\ %
    S-PEC~\cite{zhao2020noticing}                   & 1.04 & 2.18 \\ %
    DAG-Net~\cite{monti2021dag}                     & 0.77 & 1.61 \\ %
    TrajAirNet~\cite{patrikar2022trajair}           & 0.79 & 1.58 \\ %
    SocialPatteRNN-ATT~\cite{navarro2022social}     & 0.67 & 1.40 \\ %
    PECNet~\cite{mangalam2020not}                   & 0.67 & 1.14 \\ %
    MID~\cite{gu2022stochastic}                     & 0.55 & 0.87 \\ %
    Expert-Traj~\cite{zhao2021expert}               & 0.55 & 0.72 \\ %
    GooDFlight~\cite{yang2025goodflight}            & \sv{0.29} & \sv{0.39} \\ %
    \ourrow                                         & \bv{0.19} & \bv{0.26} \\ %
    \multicolumn{2}{c}{} \\  [-0.2cm]
    \multicolumn{1}{c||}{\emph{Input: 11 s}} & \multicolumn{2}{c}{111Days} \\
    \multicolumn{1}{c||}{Method} & minADE$_{5}$ & minFDE$_{5}$ \\ \hline 
    \ourrow                                         & \bv{0.29} & \bv{0.49} \\ %
    \end{tabular}
    \end{center}
    \vspace{-0.7cm}
\setlength{\tabcolsep}{4pt}
\end{table}

\begin{table}[t]
\setlength{\tabcolsep}{3pt}
    \vspace{0.17cm}
    \caption{
    Trajectory prediction results on the TartanAviation dataset~\cite{patrikar2025image}. The dataset contains additional data from the \icaoone~airport (same location as TrajAir dataset) as well as data from the towered \icaotwo~airport.
    Errors reported in \text{km}.
    }
    \vspace{-0.4cm}
    \label{tab:res_tartan}
    \begin{center}
    \begin{tabular}{l||cc|cc}
    \multicolumn{1}{c||}{\emph{Input: 11 s}} & \multicolumn{2}{c|}{\icaoone~S1} & \multicolumn{2}{c}{\icaoone~S2} \\
    \multicolumn{1}{c||}{Method}             & minADE$_{5}$ & minFDE$_{5}$ & minADE$_{5}$ & minFDE$_{5}$ \\ \hline 
    Constant Velocity                        & 1.83 & 3.91 &1.77 & 3.80 \\
    \ourrow                                  & \bv{0.36} & \bv{0.60} & \bv{0.32} & \bv{0.56} \\ %
    \multicolumn{2}{c}{} \\  [-0.2cm]
    \multicolumn{1}{c||}{\emph{Input: 11 s}} & \multicolumn{2}{c|}{\icaotwo~S1} & \multicolumn{2}{c}{\icaotwo~S2} \\
    \multicolumn{1}{c||}{Method}             & minADE$_{5}$ & minFDE$_{5}$ & minADE$_{5}$ & minFDE$_{5}$ \\ \hline 
    Constant Velocity                        & 1.77 & 3.78 & 1.86 & 3.96 \\
    \ourrow                                  & \bv{0.47} & \bv{0.72} & \bv{0.50} & \bv{0.75} \\
    \end{tabular}
    \end{center}
    \vspace{-0.6cm}
\setlength{\tabcolsep}{4pt}
\end{table}

\begin{table*}[t]
\setlength{\tabcolsep}{6pt}
    \vspace{0.17cm}
    \caption{
     Ablation study on using different decoder modules (a CVAE-based baseline and our mode query-based model) and output prediction strategies (directly predicting xyz-positions vs. predicting flight parameters).
     Furthermore, we compare different coordinate modelings: global coordinates vs. local coordinates with different types of normalization.
     Errors given in km.
    }
    \vspace{-0.4cm}
    \label{tab:abl}
    \begin{center}
    \begin{tabular}{c|cccc|c||cc}
    Decoder      & Coordinate & \multicolumn{2}{c}{Normalization} & Positional & Flight Parameter & \multicolumn{2}{c}{7Days-Avg} \\
    Architecture & System     & Positional & Angular              & Embeddings & Prediction & minADE$_{5}$ & minFDE$_{5}$ \\ \hline 
    CVAE Baseline & Local & \cmark & \cmark & \xmark & \xmark & 0.68 & 1.33 \\ %
    Mode Queries & Local & \cmark & \cmark & \xmark & \xmark & 0.52 & 0.98 \\ %
    CVAE Baseline & Global & - & - & - & - & 0.48 & 0.82 \\ %
    CVAE Baseline & Local & \cmark & \cmark & \cmark & \xmark & 0.44 & 0.80 \\ %
    Mode Queries & Global & - & - & - & - & 0.38 & 0.60 \\ %
    Mode Queries                 & Local & \cmark & \xmark & \cmark & \xmark & 0.36  & 0.60 \\ %
    Mode Queries                 & Local & \cmark & \cmark & \cmark & \xmark & \bv{0.35} & 0.60 \\ %
    Mode Queries                 & Local & \cmark & \xmark & \cmark & \cmark & 0.37 & \bv{0.58} \\ %
    \rowcolor{rcol} Mode Queries & Local & \cmark & \cmark & \cmark & \cmark & \bv{0.35} & \bv{0.58} \\ %
    \end{tabular}
    \end{center}
    \vspace{-0.6cm}
\setlength{\tabcolsep}{4pt}
\end{table*}

\begin{table}[t]
\setlength{\tabcolsep}{4pt}
    \vspace{0.17cm}
    \caption{
    Cross-dataset evaluation using the TrajAir and TartanAviation~(TA) datasets.
    *Trajectories exceeding the TrajAir dataset recording radius are filtered out.}
    \vspace{-0.4cm}
    \label{tab:cross_dataset}
    \begin{center}
    \begin{tabular}{cc||cc}
    Train Dataset & Test Dataset & minADE$_{5}$ & minFDE$_{5}$ \\ \hline 
    TrajAir 111days & TA \icaoone~S1\phantom{*} & 0.69 & 1.25 \\
    TrajAir 111days & TA \icaoone~S2\phantom{*} & 0.64 & 1.19 \\
    TrajAir 111days & TA \icaoone~S1* & 0.33 & 0.58 \\
    TrajAir 111days & TA \icaoone~S2* & 0.36 & 0.60 \\ \hline
    TA \icaoone~S1 & TrajAir 111days & 0.32 & 0.57 \\
    TA \icaoone~S2 & TrajAir 111days & 0.33 & 0.57 \\
    \end{tabular}
    \end{center}
    \vspace{-0.5cm}
\setlength{\tabcolsep}{4pt}
\end{table}

\subsection{Ablation Study}
For all ablation studies, we follow the experimental setup of TrajAirNet~\cite{patrikar2022trajair}: model input of 11\,s and prediction of $k=5$ hypothesis for the following 120\,s.
Table~\ref{tab:abl} presents an ablation study evaluating three key components of our approach: (1) decoder module, (2) input/output coordinate-frame modeling, and (3) prediction format (direct positions vs. parameterized trajectories).
First, we compare a baseline CVAE-based decoder~\cite{yan2016attribute2image}, similar to the one used in TrajAirNet~\cite{patrikar2022trajair}, against our proposed mode query-based decoder. Our decoder consistently outperforms the CVAE baseline, demonstrating that a mode query-based approach is better suited for the aircraft problem setting.
Second, we investigate different  input and output coordinate representations.
We evaluate models using global coordinates, and locally normalized coordinates with or without global positional embeddings.
Using only local motion features yields the weakest performance, since the model lacks global context.
Directly using global coordinates performs well, but the best results are obtained with local normalization combined with positional embeddings; this way the model has relative motion structure and global reference.
Finally, we compare two prediction formats: direct decoding of 3D positions versus parameterized trajectory decoding.
For long-horizon forecasting, evaluated by the minimum final displacement error (minFDE), the parameter-based approach achieves more accurate results, producing more stable trajectories.

Table~\ref{tab:cross_dataset} presents cross-dataset results, where we train on either the TrajAir~\cite{patrikar2022trajair} or TartanAviation~\cite{patrikar2025image} dataset and evaluate on the other.
Notably, TartanAviation spans a larger recording radius, whereas TrajAir restricts data to within 5$\,$km of the runway.
Training on TartanAviation and evaluating on TrajAir (\emph{111Days} split) yields strong results.
In contrast, training on TrajAir and evaluating on TartanAviation leads to issues with distant trajectories, as these are not represented in the TrajAir training data.
This is confirmed by the fact that cross-dataset performance aligns closely when evaluation is restricted to trajectory ranges present in TrajAir.

\subsection{Latency Analysis}
Table~\ref{tab:latency} compares the latency of \mn~against \mbox{TrajAirNet~\cite{patrikar2022trajair}}, to our knowledge the only related work with a public implementation.
Even without optimization techniques, \mbox{\mn}~achieves significantly lower latency across all batch sizes (\ie number of aircraft).
This performance stems from two primary architectural advantages: the direct prediction of $k$ hypotheses, which eliminates iterative decoder sampling, and a streamlined transformer- and MLP-based design.
These results demonstrate that our approach is well-suited for edge deployment, providing a viable path toward real-time conflict detection to enhance safety in GA.

\begin{table}[t]
\setlength{\tabcolsep}{4pt}
    \vspace{0.17cm}
    \caption{
    Inference latency analysis on a single NVIDIA V100 GPU across different numbers of aircraft ($B$).
    }
    \vspace{-0.4cm}
    \label{tab:latency}
    \begin{center}
    \begin{tabular}{c||ccccc}
    & \multicolumn{5}{c}{Latency in ms} \\
    Method & $B=1$ & $B=4$ & $B=8$  & $B=16$ & $B=32$ \\ \hline
    TrajAirNet~\cite{patrikar2022trajair} & 57 & 118 & 191 & 350 & 673 \\
    \ourrow                            & 11 & \phantom{0}13 & \phantom{0}14 & \phantom{0}18 & \phantom{0}17 
    \end{tabular}
    \end{center}
    \vspace{-0.6cm}
\setlength{\tabcolsep}{4pt}
\end{table}

\section{CONCLUSIONS}
We present a transformer-based trajectory prediction model that achieves state-of-the-art results on both the \mbox{TrajAir}~\cite{patrikar2022trajair} and TartanAviation~\cite{patrikar2022trajair} datasets.
By decomposing the encoding into local motion and global context, and introducing a mode query-based decoding of flight parameters, our model achieves highly accurate multi-modal predictions.
A limitation of our approach is the implicit learning of global relationships, such as runway locations, which affect generalization to other airports without finetuning.
This is primarily due to the limited availability of data from diverse non-towered airspaces, which constrains broader training and evaluation.
Currently, we do not use additional information, such as weather conditions. However, future work can easily extend our model design to encompass such features.

\bibliography{chapters/_ref}

\end{document}